\documentclass[journal]{IEEEtran}
\usepackage{blindtext}
\usepackage{graphicx}
\usepackage{amsmath}
\usepackage{xfrac}
\usepackage{subcaption}

\ifCLASSINFOpdf
\else
\fi
\begin{document}
%

\title{An Empirical Evaluation of Deep Learning on Highway Driving}

%
%
%

\author{
\IEEEauthorblockN{Brody~Huval\IEEEauthorrefmark{1},~Tao~Wang\IEEEauthorrefmark{1}, 
                  Sameep~Tandon\IEEEauthorrefmark{1},~Jeff~Kiske\IEEEauthorrefmark{1},
                  Will~Song\IEEEauthorrefmark{1},~Joel~Pazhayampallil\IEEEauthorrefmark{1},
                  Mykhaylo~Andriluka\IEEEauthorrefmark{1},~Pranav~Rajpurkar\IEEEauthorrefmark{1},
                  ~Toki~Migimatsu\IEEEauthorrefmark{1},~Royce~Cheng-Yue\IEEEauthorrefmark{2},
                  Fernando~Mujica\IEEEauthorrefmark{3},~Adam~Coates\IEEEauthorrefmark{4},
                  Andrew~Y.~Ng\IEEEauthorrefmark{1}} \\
\IEEEauthorblockA{\IEEEauthorrefmark{1}Stanford University}~
\IEEEauthorblockA{\IEEEauthorrefmark{2}Twitter}~
\IEEEauthorblockA{\IEEEauthorrefmark{3}Texas Instruments}~
\IEEEauthorblockA{\IEEEauthorrefmark{4}Baidu Research} \\
}

\maketitle

\begin{abstract}
\boldmath
Numerous groups have applied a variety of deep learning techniques to computer vision problems in highway perception scenarios. In this paper, we presented a number of empirical evaluations of recent deep learning advances. Computer vision, combined with deep learning, has the potential to bring about a relatively inexpensive, robust solution to autonomous driving. To prepare deep learning for industry uptake and practical applications, neural networks will require large data sets that represent all possible driving environments and scenarios. We collect a large data set of highway data and apply deep learning and computer vision algorithms to problems such as car and lane detection. We show how existing convolutional neural networks (CNNs) can be used to perform lane and vehicle detection while running at frame rates required for a real-time system. Our results lend credence to the hypothesis that deep learning holds promise for autonomous driving. 

\end{abstract}


%
\IEEEpeerreviewmaketitle

\section{Introduction}
Since the DARPA Grand Challenges for autonomous vehicles, there has been an explosion in applications and research for self-driving cars. Among the different environments for self-driving cars, highway and urban roads are on opposite ends of the spectrum. In general, highways tend to be more predictable and orderly, with road surfaces typically well-maintained and lanes well-marked. In contrast, residential or urban driving environments feature a much higher degree of unpredictability with many generic objects, inconsistent lane-markings, and elaborate traffic flow patterns. The relative regularity and structure of highways has facilitated some of the first practical applications of autonomous driving technology. Many automakers have begun pursuing highway auto-pilot solutions designed to mitigate driver stress and fatigue and to provide additional safety features; for example, certain advanced-driver assistance systems (ADAS) can both keep cars within their lane and perform front-view car detection. Currently, the human drivers retain liability and, as such, must keep their hands on the steering wheel and prepare to control the vehicle in the event of any unexpected obstacle or catastrophic incident. Financial considerations contribute to a substantial performance gap between commercially available auto-pilot systems and fully self-driving cars developed by Google and others. Namely, today's self-driving cars are equipped with expensive but critical sensors, such as LIDAR, radar and high-precision GPS coupled with highly detailed maps.

\begin{figure}[t]
  \centering
    \includegraphics[width=3.4in]{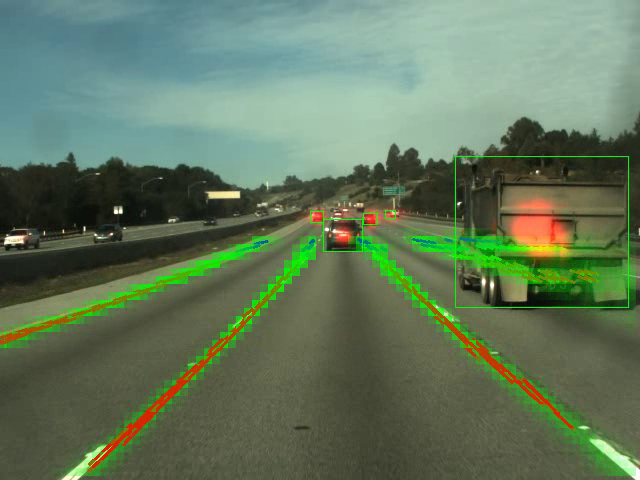}
 \caption{Sample output from our neural network capable of lane and vehicle detection.}
 \label{fig:car-lanes}
\end{figure}
In today's production-grade autonomous vehicles, critical sensors include radar, sonar, and cameras. Long-range vehicle detection typically requires radar, while nearby car detection can be solved with sonar. Computer vision can play an important a role in lane detection as well as redundant object detection at moderate distances. Radar works reasonably well for detecting vehicles, but has difficulty distinguishing between different metal objects and thus can register false positives on objects such as tin cans. Also, radar provides little orientation information and has a higher variance on the lateral position of objects, making the localization difficult on sharp bends. The utility of sonar is both compromised at high speeds and, even at slow speeds, is limited to a working distance of about $2$ meters. Compared to sonar and radar, cameras generate a richer set of features at a fraction of the cost. By advancing computer vision, cameras could serve as a reliable redundant sensor for autonomous driving. Despite its potential, computer vision has yet to assume a significant role in today's self-driving cars. Classic computer vision techniques simply have not provided the robustness required for production grade automotives; these techniques require intensive hand engineering, road modeling, and special case handling. Considering the seemingly infinite number of specific driving situations, environments, and unexpected obstacles, the task of scaling classic computer vision to robust, human-level performance would prove monumental and is likely to be unrealistic.

Deep learning, or neural networks, represents an alternative approach to computer vision. It shows considerable promise as a solution to the shortcomings of classic computer vision. Recent progress in the field has advanced the feasibility of deep learning applications to solve complex, real-world  problems; industry has responded by increasing uptake of such technology. Deep learning is data centric, requiring heavy computation but minimal hand-engineering. In the last few years, an increase in available storage and compute capabilities have enabled deep learning to achieve success in supervised perception tasks, such as image detection. A neural network, after training for days or even weeks on a large data set, can be capable of inference in real-time with a model size that is no larger than a few hundred MB~\cite{krizhevsky-2012}. State-of-the-art neural networks for computer vision require very large training sets coupled with extensive networks capable of modeling such immense volumes of data. For example, the ILSRVC data-set, where neural networks achieve top results, contains $1.2$ million images in over $1000$ categories. 

By using expensive existing sensors which are currently used for self-driving applications, such as LIDAR and mm-accurate GPS, and calibrating them with cameras, we can create a video data set containing labeled lane-markings and annotated vehicles with location and relative speed. By building a labeled data set in all types of driving situations (rain, snow, night, day, etc.), we can evaluate neural networks on this data to determine if it is robust in every driving environment and situation for which we have training data. 

In this paper, we detail empirical evaluation on the data set we collect. In addition, we explain the neural network that we applied for detecting lanes and cars, as shown in Figure~\ref{fig:car-lanes}.

\section{Related Work}
Recently, computer vision has been expected to player a larger role within autonomous driving. However, due to its history of relatively low precision, it is typically used in conjunction with either other sensors or other road models \cite{cho-2014,held-2012,carafii-2012,jazayeri-2011}. Cho \textit{et al.} \cite{cho-2014} uses multiple sensors, such as LIDAR, radar, and computer vision for object detection. They then fuse these sensors together in a Kalman filter using motion models on the objects. Held \textit{et al.} \cite{held-2012}, uses only a deformable parts based model on images to get the detections, then uses road models to filter out false positives. Carafii \textit{et al.} \cite{carafii-2012} uses a WaldBoost detector along with a tracker to generate pixel space detections in real time. Jazayeri \textit{et al.} \cite{jazayeri-2011} relies on temporal information of features for detection, and then filters out false positives with a front-view motion model.

In contrast to these object detectors, we do not use any road or motion-based models; instead we rely only on the robustness of a neural network to make reasonable predictions. In addition, we currently do not rely on any temporal features, and the detector operates independently on single frames from a monocular camera. To make up for the lack of other sensors, which estimate object depth, we train the neural network to predict depth based on labels extracted from radar returns. Although the model only predicts a single depth value for each object, Eigen \textit{et al.} have shown how a neural network can predict entire depth maps from single images~\cite{eigen-2014}. The network we train likely learns some model of the road for object detection and depth predictions, but it is never explicitly engineered and instead learns from the annotations alone. 

Before the wide spread adoption of Convolutional Neural Networks (CNNs) within computer vision, deformable parts based models were the most successful methods for detection~\cite{felzenszwalb-2010}. After the popular CNN model AlexNet~\cite{krizhevsky-2012} was proposed, state-of-the-art detection shifted towards CNNs for feature extraction~\cite{sermanet-2013,szegedy-2014,szegedy-2013,girshick-2014}. Girshick \textit{et al.} developed R-CNN, a two part system which used Selective Search~\cite{uijlings-2013} to propose regions and AlexNet to classify them. R-CNN achieved state-of-the-art on Pascal by a large margin; however, due to its nearly $1000$ classification queries and inefficient re-use of convolutions, it remains impractical for real-time implementations. Szegedy \textit{et al.} presented a more scalable alternative to R-CNN, that relies on the CNN to propose higher quality regions compared to Selective Search. This reduces the number of region proposals down to as low as $79$ while keeping the mAP competitive with Selective Search. An even faster approach to image detection called Overfeat was presented by Sermanet~\textit{et al.}~\cite{sermanet-2013}. By using a regular pattern of ``region proposals'', Overfeat can efficiently reuse convolution computations from each layer, requiring only a single forward pass for inference.  
 
For our empirical evaluation, we use a straight-forward application of Overfeat, due to its efficiencies, and combine this with labels similar to the ones proposed by Szegedy~\textit{et al.}. We describe the model and similarities in the next section. 
 
\section{Real Time Vehicle Detection}
Convolutional Neural Networks (CNNs) have had the largest success in image recognition in the past 3 years~\cite{krizhevsky-2012,szegedy-2014-inception,he-2015,simonyan-2014}. From these image recognition systems, a number of detection networks were adapted, leading to further advances in image detection. While the improvements have been staggering, not much consideration had been given to the real-time detection performance required for some applications. In this paper, we present a detection system capable of operating at greater than $10$Hz using nothing but a laptop GPU. Due to the requirements of highway driving, we need to ensure that the system used can detect cars more than $100$m away and can operate at speeds greater than $10$Hz; this distance requires higher image resolutions than is typically used, and in our case is $640 \times 480$. We use the Overfeat CNN detector, which is very scalable, and simulates a sliding window detector in a single forward pass in the network by efficiently reusing convolutional results on each layer \cite{sermanet-2013}. Other detection systems, such as R-CNN, rely on selecting as many as $1000$ candidate windows, where each is evaluated independently and does not reuse convolutional results. 

In our implementation, we make a few minor modifications to Overfeat's labels in order to handle occlusions of cars, predictions of lanes, and accelerate performance during inference. We will first provide a brief overview of the original implementation and will then address the modifications. Overfeat converts an image recognition CNN into a ``sliding window'' detector by providing a larger resolution image and transforming the fully connected layers into convolutional layers. Then, after converting the fully connected layer, which would have produced a single final feature vector, to a convolutional layer, a grid of final feature vectors is produced. Each of the resulting feature vectors represents a slightly different context view location within the original pixel space. To determine the stride of this window in pixel space, it is possible to simply multiply the strides on each convolutional or pool layer together. The network we used has a stride size of $32$ pixels. Each final feature vector in this grid can predict the presence of an object; once an object is detected, those same features are then used to predict a \textit{single} bounding box through regression. The classifier will predict ``no-object'' if it can not discern any part of an object within its \textit{entire} input view. This causes large ambiguities for the classifier, which can only predict a single object, as two different objects could can easily appear in the context view of the final feature vector, which is typically larger than $50\%$ of the input image resolution. 

The network we used has a context view of $355 \times 355$ pixels in size. To ensure that all objects in the image are classified at least once, many different context views are taken of the image by using skip gram kernels to reduce the stride of the context views and by using up to four different scales of the input image. The classifier is then trained to activate when an object appears anywhere within its entire context view. In the original Overfeat paper, this results in $1575$ different context views (or final feature vectors), where each one is likely to become active (create a bounding box).

This creates two problems for our empirical evaluation. Due to the L2 loss between the predicted bounding box and actual bounding proposed by Sermanet \textit{et al.}, the ambiguity of having two valid bounding box locations to predict when two objects appear, is incorrectly handled by the network by predicting a box in the center of the two objects to minimize its expected loss. These boxes tend to cause a problem for the bounding box merging algorithm, which incorrectly decides that there must be a third object between the two ground truth objects. This could cause problems for an ADAS system which falsely believes there is a car where there is not, and emergency breaking is falsely applied. In addition, the merging algorithm, used only during inference, operates in $O(n^2)$ where $n$ is the number of bounding boxes proposed. Because the bounding box merging is not as easily parallelizable as the CNN, this merging may become the bottleneck of a real-time system in the case of an inefficient implementation or too many predicted bounding boxes.  

\begin{figure}[tb]
  \centering
    \includegraphics[width=3.4in]{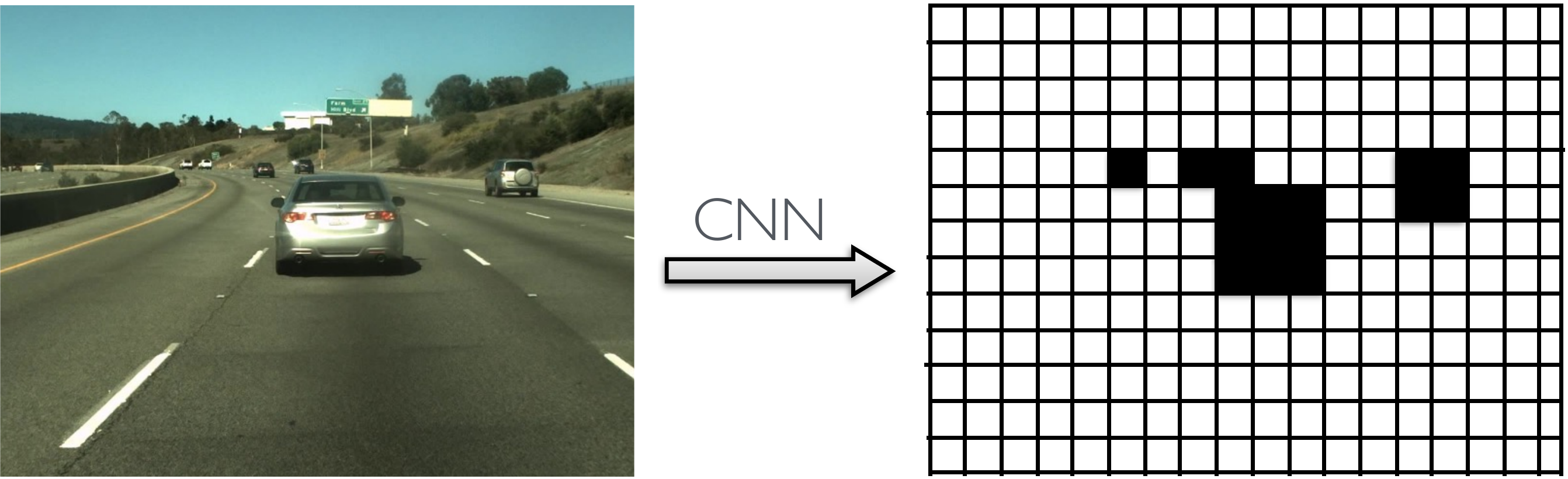}
 \caption{mask detector}
 \label{fig:mask-detector}
\end{figure}

In our evaluations, we use a mask detector as described in Szegedy \textit{et al.}~\cite{szegedy-2013} to improve some of the issues with Overfeat as described above. Szegedy \textit{et al.} proposes a CNN that takes an image as input and outputs an object mask through regression, highlighting the object location. The idea of a mask detector is shown in Fig~\ref{fig:mask-detector}. To distinguish multiple nearby objects, different part-detectors output object masks, from which bounding boxes are then extracted. The detector they propose must take many crops of the image, and run multiple CNNs for each part on every crop. Their resulting implementation takes roughly $5$-$6$ seconds per frame per class using a 12-core machine, which would be too slow for our application. 

\begin{figure*}[tb]
  \centering
    \includegraphics[width=7.1in]{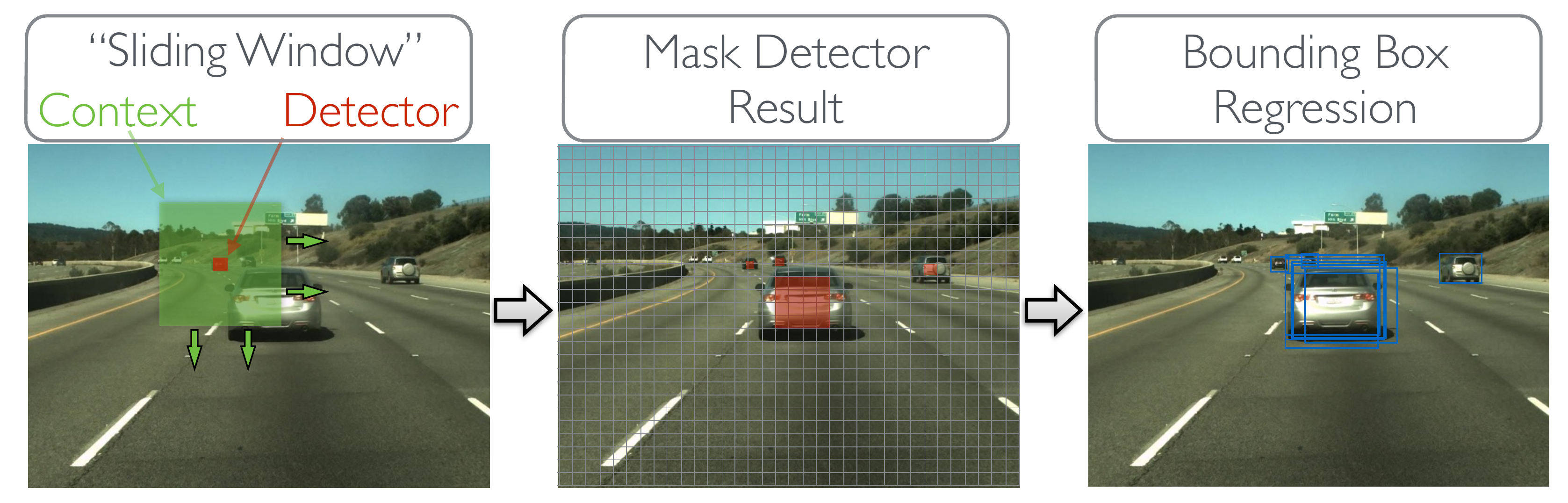}
 \caption{overfeat-mask}
 \label{fig:overfeat-mask}
\end{figure*}

We combine these ideas by using the efficient ``sliding window'' detector of Overfeat to produce an object mask and perform bounding box regression. This is shown in Fig~\ref{fig:overfeat-mask}. In this implementation, we use a single image resolution of $640 \times 480$ with no skip gram kernels. To help the ambiguity problem, and reduce the number of bounding boxes predicted, we alter the detector on the top layer to only activate within a $4 \times 4$ pixel region at the center of its context view, as shown in the first box in Fig~\ref{fig:overfeat-mask}. Because it's highly unlikely that any two different object's bounding boxes appear in a $4 \times 4$ pixel region, compared to the entire context view with Overfeat, the bounding box regressor will no longer have to arbitrarily choose between two valid objects in its context view. In addition, because the requirement for the detector to fire is stricter, this produces many fewer bounding boxes which greatly reduces our run-time performance during inference. 

Although these changes helped, ambiguity was still a common problem on the border of bounding boxes in the cases of occlusion. This ambiguity results in a false bounding box being predicted between the two ground truth bounding boxes. To fix this problem, the bounding boxes were first shrunk by $75\%$ before creating the detection mask label. This added the additional requirement that the center $4 \times 4$-pixel region of the detector window had to be within the center region of the object before activating. The bounding box regressor however, still predicts the original bounding box before shrinking. This also further reduces the number of active bounding boxes as input to our merging algorithm. We also found that switching from L2 to L1 loss on the bounding box regressions results in better performance. To merge the bounding boxes together, we used OpenCV's efficient implementation of \texttt{groupRectangles}, which clusters the bounding boxes based on a similarity metric in $O(n^2)$~\cite{opencv}. 

The lower layers of our CNN we use for feature extraction is similar to the one proposed by Krizhevsky \textit{et al.}~\cite{krizhevsky-2012}. Our modifications to the network occurs on the dense layers which are converted to convolution, as described in Sermanet \textit{et al.}~\cite{sermanet-2013}. When using our larger image sizes of $640 \times 480$ this changes the previous final feature response maps of size $1\times 1\times 4096$ to $20 \times 15 \times 4096$. As stated earlier, each of these feature vectors sees a context region of $355 \times 355$ pixels, and the stride between them is $32 \times 32$ pixels; however, we want each making predictions at a resolution of $4 \times 4$ pixels, which would leave gaps in our input image. To fix this, we use each $4096$ feature as input to $64$ softmax classifiers, which are arranged in an $8 \times 8$ grid each predicting if an object is within a different $4 \times 4$ pixel region. This allows for the $4096$ feature vector to cover the full stride size of $32 \times 32$ pixels; the end result is a grid mask detector of size $160 \times 120$ where each element is $4 \times 4$ pixels which covers the entire input image of size $640 \times 480$. 

\subsection{Lane Detection}
\begin{figure}[tb]
  \centering
    \includegraphics[width=3.4in]{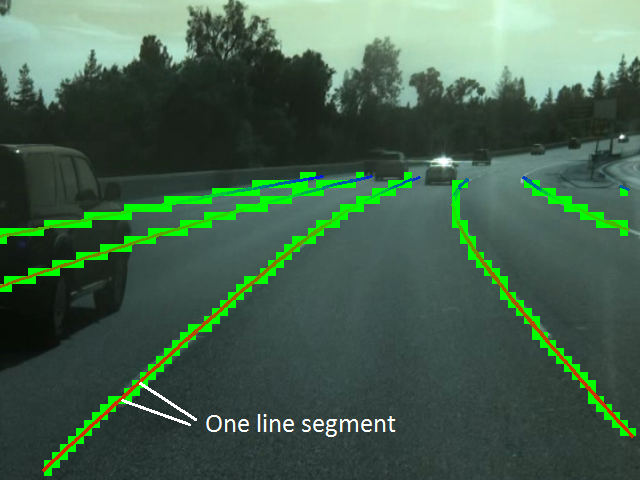}
 \caption{Example of lane boundary ground truth}
 \label{fig:lane-gt}
\end{figure}

The CNN used for vehicle detection can be easily extended for lane boundary detection by adding an additional class. Whereas the regression for the vehicle class predicts a five dimensional value (four for the bounding box and one for depth), the lane regression predicts six dimensions. Similar to the vehicle detector, the first four dimensions indicate the two end points of a local line segment of the lane boundary. The remaining two dimensions 
indicate the depth of the endpoints with respect to the camera. Fig~\ref{fig:lane-gt} visualizes the lane boundary ground truth label overlaid on an example image. The green tiles indicate locations where the detector is trained to fire, and the line segments represented by the regression labels are explicitly drawn. The line segments have their ends connected to form continuous splines. The depth of the line segments are color-coded such that the closest segments are red and the furthest ones are blue. Due to our data collection methods for lane labels, we are able to obtain ground truth in spite of objects that occlude them. This forces the neural network to learn more than a simple paint detector, and must use context to predict lanes where there are occlusions.

\begin{figure}[tb]
  \centering
    \includegraphics[width=3.4in]{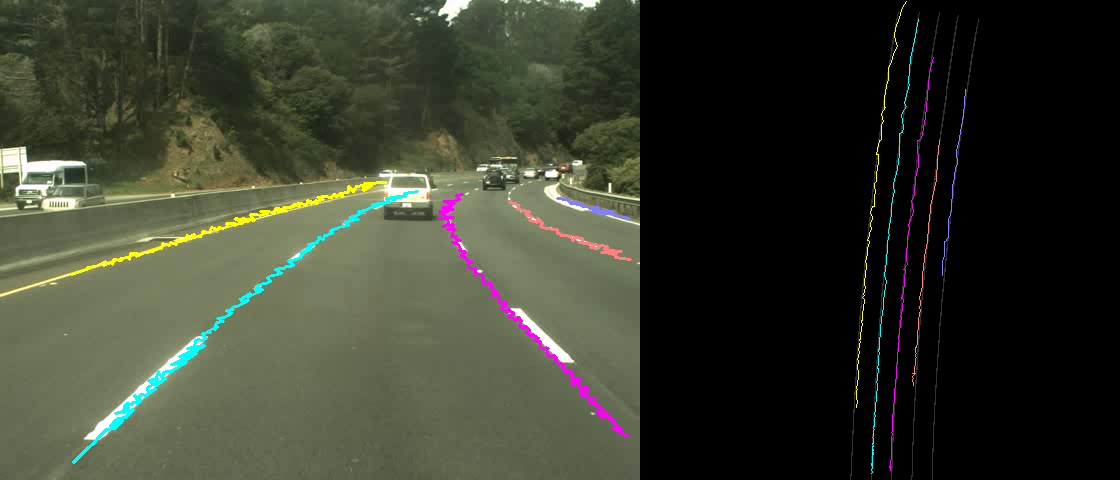}
 \caption{Example output of lane detector after DBSCAN clustering}
 \label{fig:lane-out-dbscan}
\end{figure}

Similar to the vehicle detector, we use L1 loss to train the regressor. We use mini-batch stochastic gradient descent for optimization. The learning rate is controlled by a variant of the momentum scheduler~\cite{sutskever-2013}. To obtain semantic lane information, we use DBSCAN to cluster the line segments into lanes. Fig~\ref{fig:lane-out-dbscan} shows our lane predictions after DBSCAN clustering. Different lanes are represented by different colors. Since our regressor outputs depths as well, we can predict the lane shapes in 3D using inverse camera perspective mapping.
 

\section{Experimental Setup}

\subsection{Data Collection}
Our Research Vehicle is a 2014 Infiniti Q50. The car currently uses the following sensors: 6x Point Grey Flea3 cameras, 1x Velodyne HDL32E lidar, and 1x Novatel SPAN-SE Receiver. We also have access to the Q50 built-in Continental mid-range radar system. The sensors are connected to a Linux PC with a Core i7-4770k processor. 

Once the raw videos are collected, we annotate the 3D locations for vehicles and lanes as well as the relative speed of all the vehicles. To get vehicle annotations, we follow the conventional approach of using Amazon Mechanical Turk to get accurate bounding box locations within pixel space. Then, we match bounding boxes and radar returns to obtain the distance and relative speed of the vehicles. 

Unlike vehicles that can be annotated with bounding boxes, highway lane borders often need to be annotated as curves of various shapes. This makes frame-level labelling not only tedious and inefficient, but also prone to human errors. Fortunately, lane markings can be considered as “static” objects that do not change their geolocations very often. We follow the process descried in~\cite{levinson-2011} to create LIDAR maps of the environment using the Velodyne and GNSS systems. Using these maps, labeling is straight forward. First, we filter the 3D point clouds based on lidar return intensity and position to obtain the left and right boundaries of the ego-lane. Then, we replicate the left and right ego-lane boundaries to obtain initial guesses for all the lane boundaries. A human annotator inspects the generated lane boundaries and makes appropriate corrections using our 3D labelling tool. For completeness, we describe each of these steps in details.
 
\subsubsection{Ego-lane boundary generation}
Since we do not change lanes during data collection drives, the GPS trajectory of our research vehicle already gives a decent estimate of the shape of the road. We can then easily locate the ego-lane boundaries using a few heuristic filters. Noting that lane boundaries on highways are usually marked with retro-reflective materials, we first filter out low-reflectivity surfaces such as asphalt in our 3D point cloud maps and only consider points with high enough laser return intensities. We then filter out other reflective surfaces such as cars and traffic signs by only considering points whose heights are close enough the ground plane.  Lastly, assuming our car drives close to the center of the lane, we filter out ground paint other than the ego-lane boundaries, such as other lane boundaries, car-pool or directional signs, by only considering markings whose absolute lateral distances from the car are smaller than 2.2 meters and greater than 1.4 meters. We can also distinguish the left boundary from the right one using the sign of the lateral distance. After obtaining the points in the left and right boundaries, we fit a piecewise linear curve similar to the GPS trajectory to each boundary.
 
\subsubsection{Semi-automatic generation of multiple lane boundaries}
We observe that the width of lanes during a single data collection run stays constant most of the time, with occasional exceptions such as merges and splits. Therefore, if we predefine the number of lanes to the left and right of the car for a single run, we can make a good initial guess of all the lane boundaries by shifting the auto-generated ego-lane boundaries laterally by multiples of the lane width. We will then rely on human annotators to fix the exception cases.
 
\subsection{Data Set}
\begin{figure}[tb]
  \centering
    \includegraphics[width=3.4in]{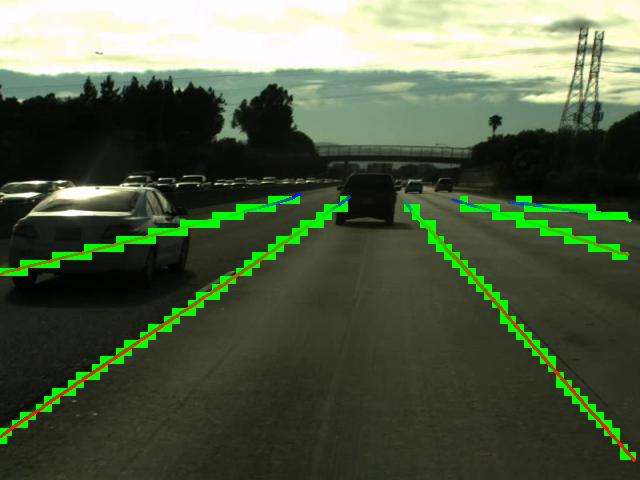}
 \caption{Image after perspective distortion}
 \label{fig:lane-gt-distort}
\end{figure}

At the time of this writing our annotated data-set consists of $14$ days of driving in the San Francisco Bay Area during the months of April-June for a few hours each day. The vehicle annotated data is sampled at $\sfrac{1}{3} \text{Hz}$ and contains nearly $17$ thousand frames with $140$ thousand bounding boxes. The lane annotated data is sampled at $5\text{Hz}$ and contains over $616$ thousand frames. During training, translation and 7 different perspective distortions are applied to the raw data sets. Fig~\ref{fig:lane-gt-distort} shows an example image after perspective distortions are applied. Note that we apply the same perspective distortion to the ground truth labels so that they match correctly with the distorted image.

\subsection{Results}
The detection network used is capable of running at $44$Hz using a desktop PC equipped with a GTX 780 Ti. When using a mobile GPU, such as the Tegra K1, we were capable of running the network at $2.5$Hz, and would expect the system to run at $5$Hz using the Nvidia PX1 chipset. 

Our lane detection test set consists of 22 video clips collected using both left and right cameras during 11 different data collection runs, which correspond to about 50 minutes of driving. We evaluate detection results for four lane boundaries, namely, the left and right boundaries of the ego lane, plus the outer boundaries of the two adjacent lanes. For each of these lane boundaries, we further break down the evaluation by longitudinal distances, which range from 15 to 80 meters ahead of the car, spaced by 5 meters. Thus, there are at maximum $4\times14=56$ positions at which we evaluate the detection results. We pair up the prediction and ground truth points at each of these locations using greedy nearest neighbor matching. True positives, false positives and false negatives are accumulated at every evaluation location in a standard way: A true positive is counted when the matched prediction and ground truth differ by less than 0.5 meter. If the matched prediction and ground truth differ by more than 0.5 meter, both false positive and false negative counts are incremented.

\begin{figure}[tb]
  \centering
    \includegraphics[width=3.4in]{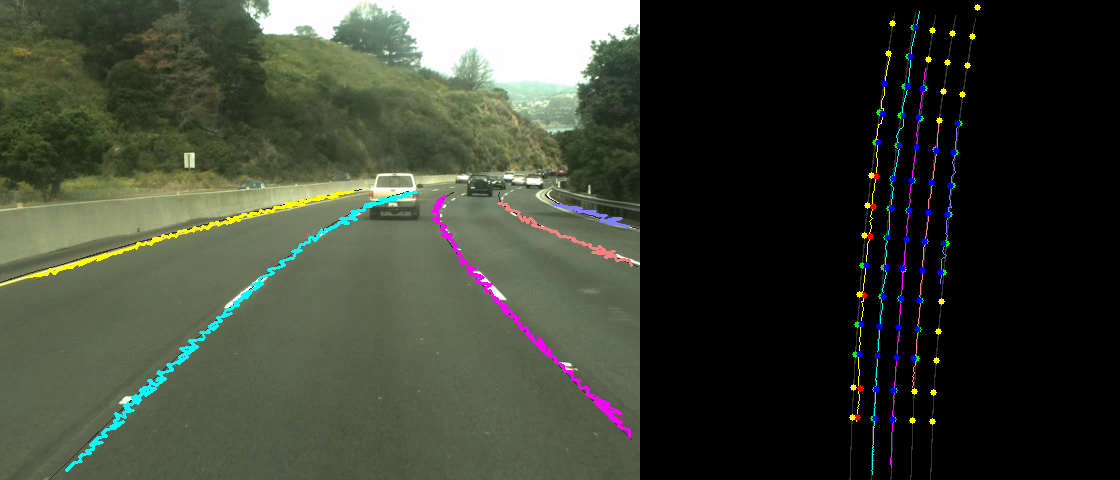}
 \caption{Left: lane prediction on test image. Right: Lane detection evaluated in 3D}
 \label{fig:lane-output-eval}
\end{figure}

\begin{figure}[tb]
  \centering
    \begin{subfigure}[b]{1.6in}
      \includegraphics[width=1.6in]{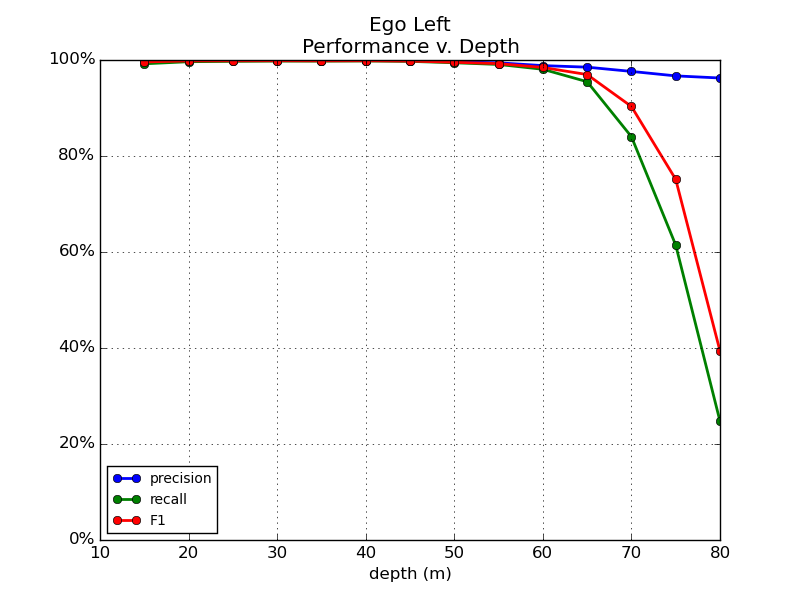}
      \caption{}
      \label{fig:lane-eval-left0}
    \end{subfigure}%
    \begin{subfigure}[b]{1.6in}
      \includegraphics[width=1.6in]{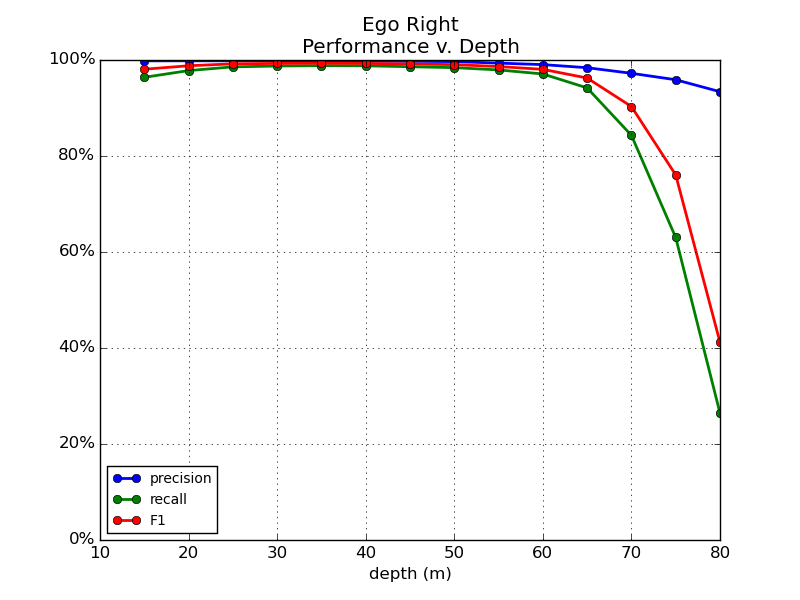}
      \caption{}
      \label{fig:lane-eval-right0}
    \end{subfigure}%
    
    \begin{subfigure}[b]{1.6in}
      \includegraphics[width=1.6in]{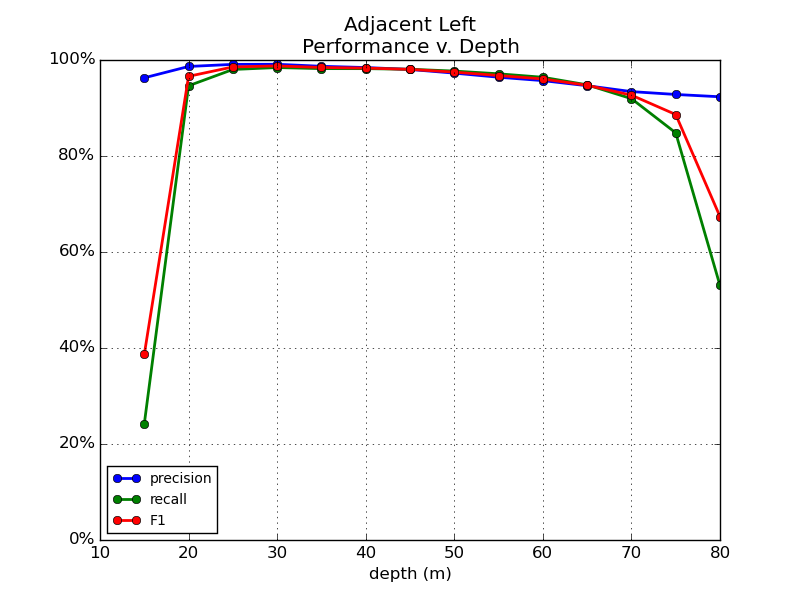}
      \caption{}
      \label{fig:lane-eval-left1}
    \end{subfigure}%
    \begin{subfigure}[b]{1.6in}
      \includegraphics[width=1.6in]{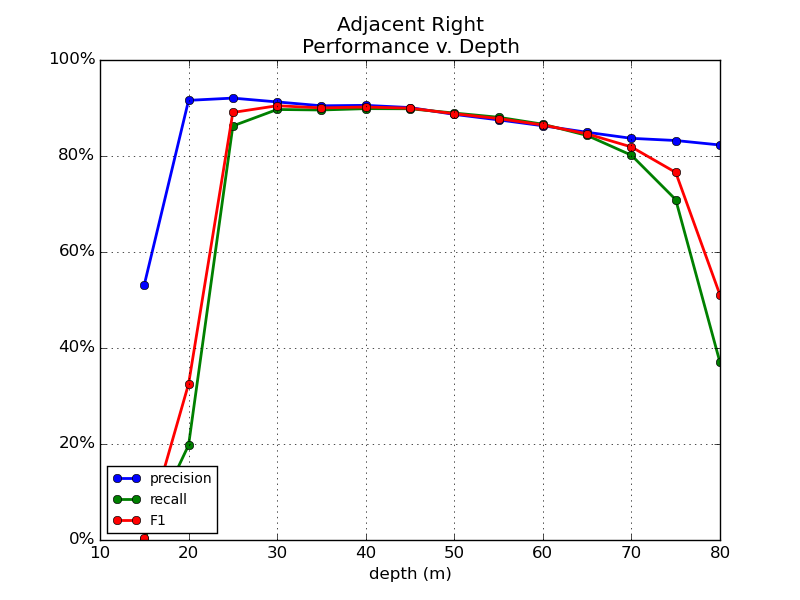}
      \caption{}
      \label{fig:lane-eval-right1}
    \end{subfigure}%
 \caption{Lane detection results on different lateral lanes. (a) Ego-lane left border. (b) Ego-lane right border. (c) Left adjacent lane left border. (d) Right adjacent lane right border.}
 \label{fig:lane-eval-plot}
\end{figure}

Fig~\ref{fig:lane-output-eval} shows a visualization of this evaluation method on one image. The blue dots are true positives. The red dots are false positives, and the yellow ones are false negatives. Fig~\ref{fig:lane-eval-plot} shows the aggregated precision, recall and F1 score on all test videos. For the ego-lane boundaries, we obtain $100\%$ F1 score up to 50 meters. Recall starts to drop fast beyond 65 meters, mainly because the resolution of the image cannot capture the width of the lane markings at that distance. For the adjacent lanes, recall is low for the nearest point because it is outside the field of view of the camera.

\begin{figure}[tb]
  \centering
    \includegraphics[width=3.4in]{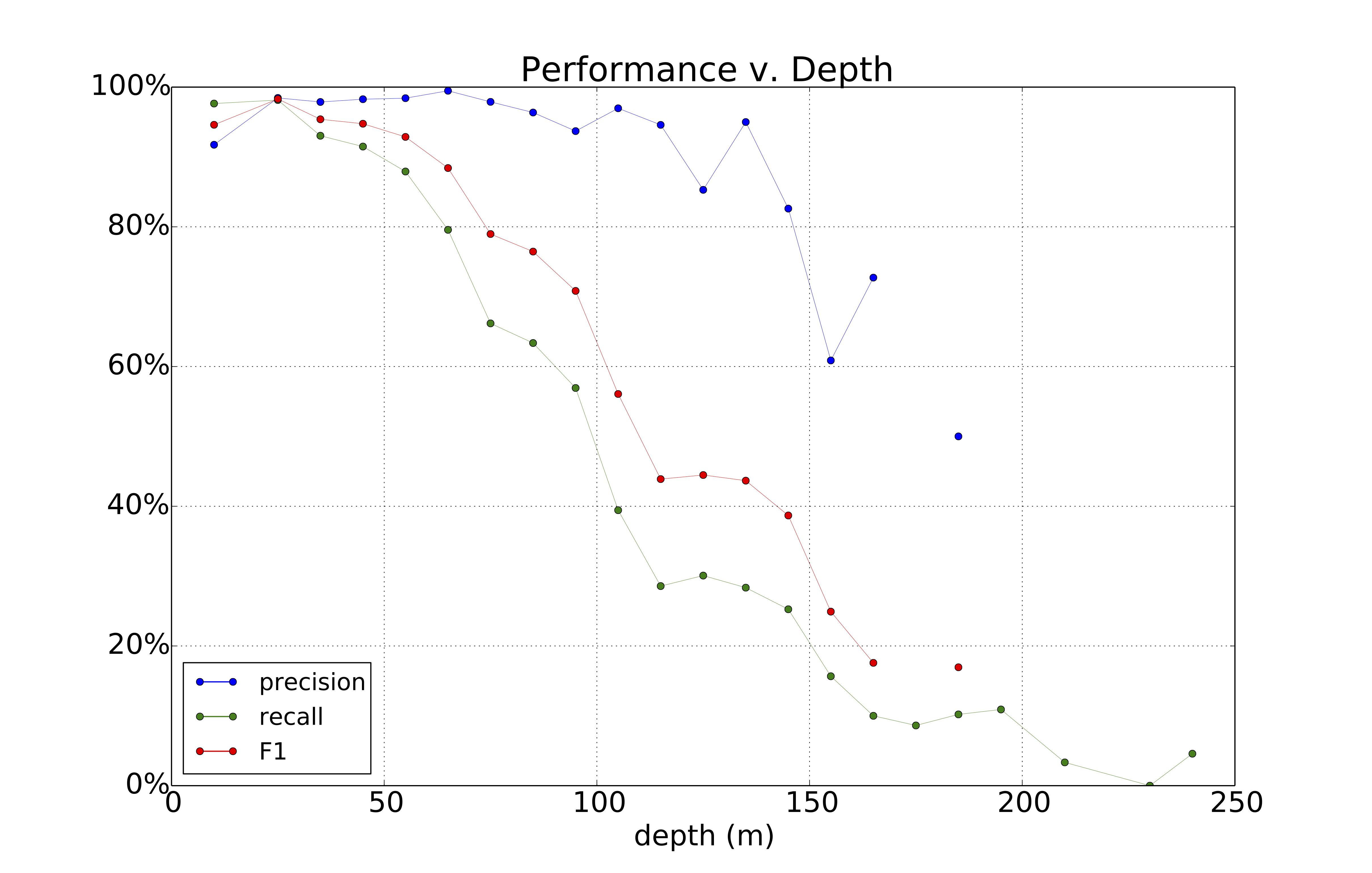}
 \caption{Car Detector Bounding Box Performance}
 \label{fig:car-bb-error}
\end{figure}

\begin{figure}[tb]
        \centering
        \begin{subfigure}[b]{1.7in}
                \includegraphics[width=1.7in]{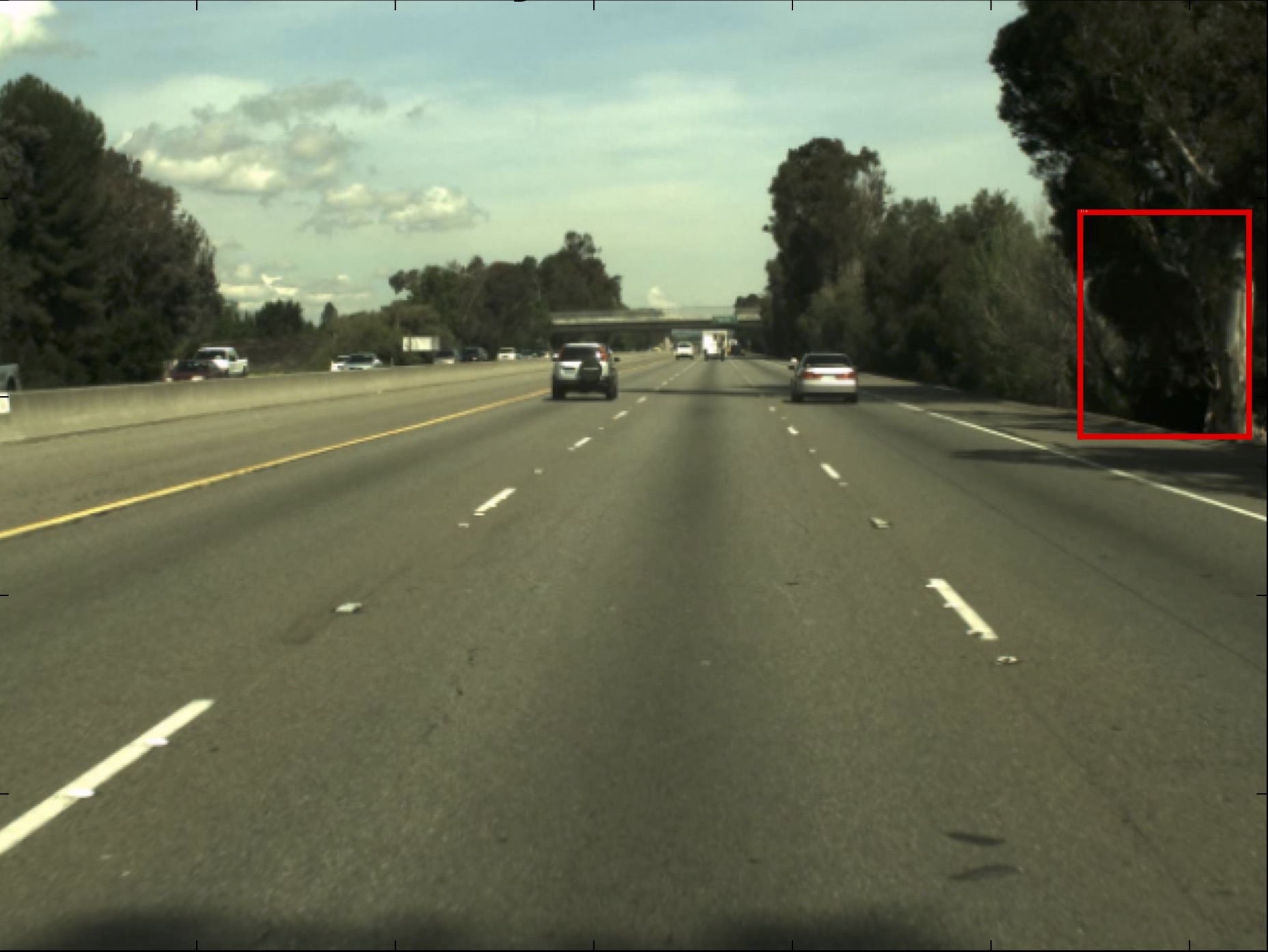}
                \caption{FP: tree}
                \label{fig:car-tree}
        \end{subfigure}%
        \begin{subfigure}[b]{1.7in}
                \includegraphics[width=1.7in]{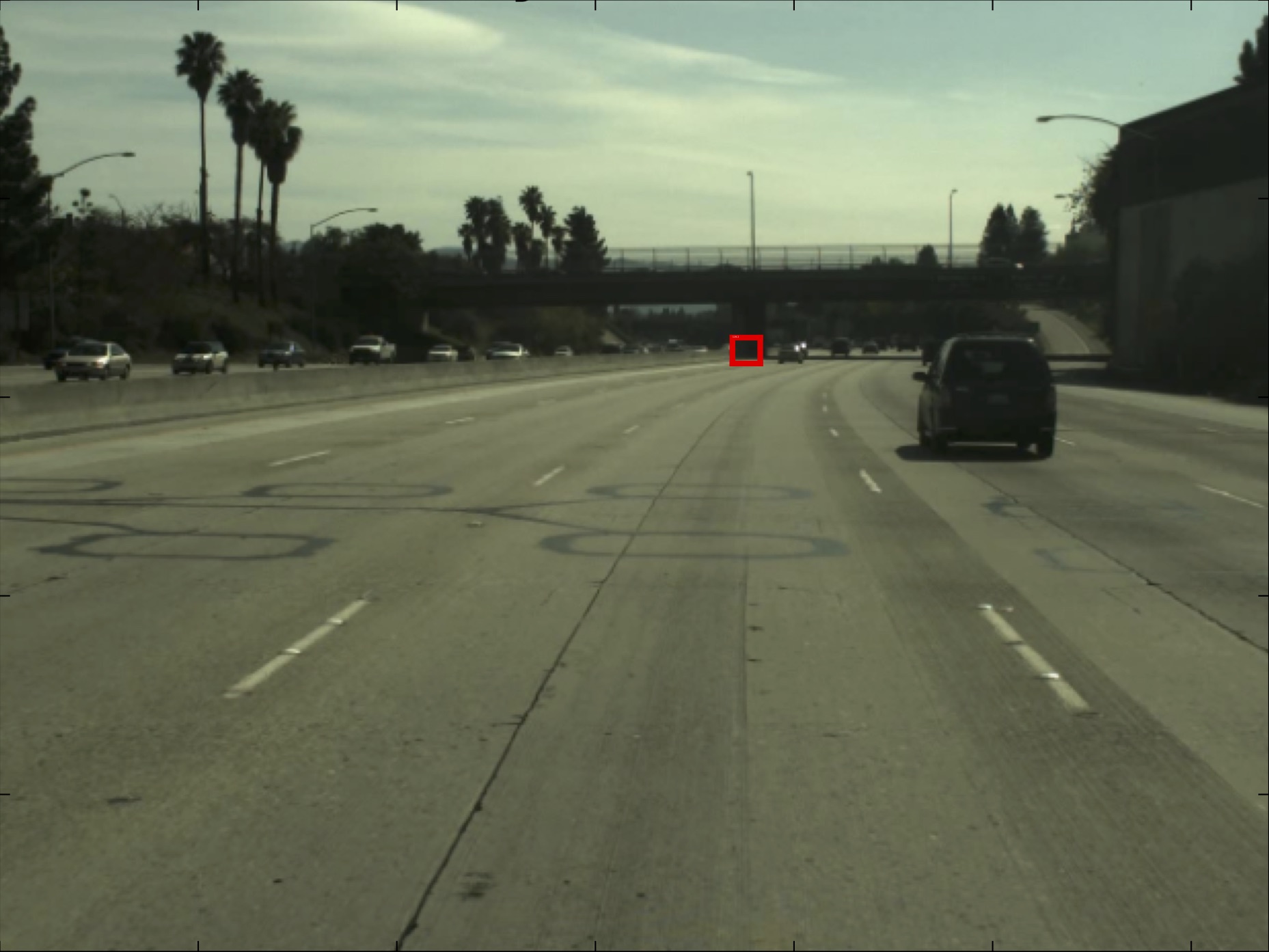}
                \caption{FP: overpass}
                \label{fig:car-overpass}
        \end{subfigure}
        \caption{Vehicle False Positives}\label{fig:car-fp}
\end{figure}

The vehicle detection test set consists of 13 video clips collected from a single day, which corresponds to 1 hour and 30 mins of driving. The accuracy of the vehicle bounding box predictions were measured using Intersection Over Union (IOU) against the ground truth boxes from Amazon Mechanical Turk (AMT). A bounding box prediction matched with ground truth if IOU$\geq0.5$. The performance of our car detection as a function of depth can be seen in Fig~\ref{fig:car-bb-error}. Nearby false positives can cause the largest problems for ADAS systems which could cause the system to needlessly apply the brakes. In our system, we found overpasses and shading effects to cause the largest problems. Two examples of these situations are shown in Fig~\ref{fig:car-fp}.

\begin{figure}[tb]
  \centering
    \includegraphics[width=3.4in]{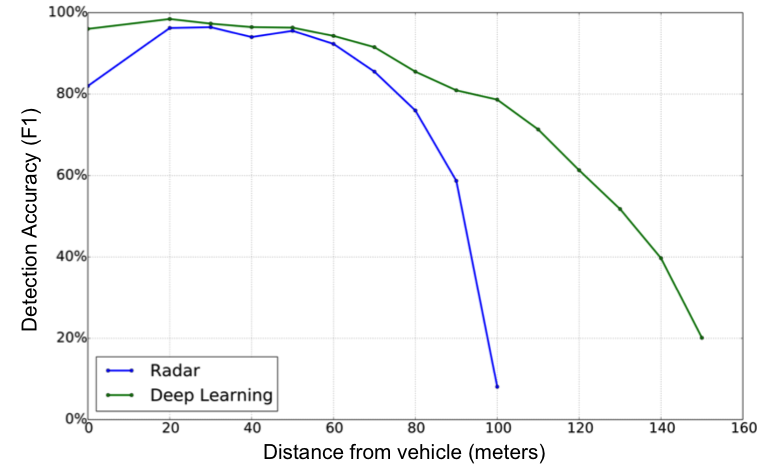}
 \caption{Radar Comparison to Vehicle Detector}
 \label{fig:nn-v-radar}
\end{figure}

As a baseline to our car detector, we compared the detection results to the Continental mid-range radar within our data collection vehicle. While matching radar returns to ground truth bounding boxes, we found that although radar had nearly $100\%$ precision, false positives were being introduced through errors in radar/camera calibration. Therefore, to ensure a fair comparison we matched every radar return to a ground truth bounding box even if IOU$<0.5$, giving our radar returns $100\%$ precision. This comparison is shown in Fig~\ref{fig:nn-v-radar}, the F1 score for radar is simply the recall.

\begin{figure}[tb]
  \centering
    \includegraphics[width=3.4in]{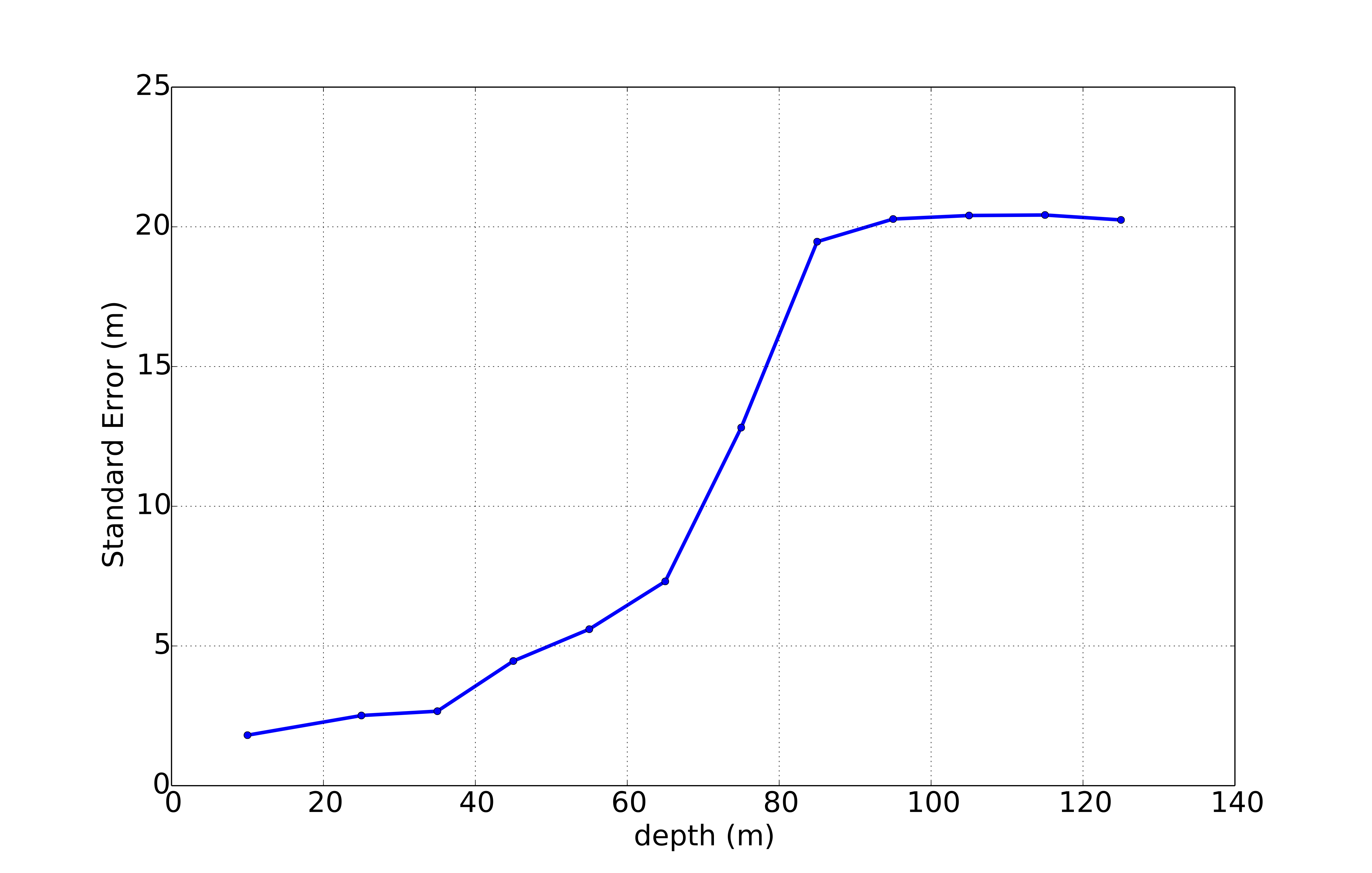}
 \caption{Car Detector Depth Performance}
 \label{fig:car-depth-error}
\end{figure}

In addition to the bounding box locations, we measured the accuracy of the predicted depth by using radar returns as ground truth. The standard error in the depth predictions as a function of depth can be seen in Fig~\ref{fig:car-depth-error}.

For a qualitative review of the detection system, we have uploaded a $1.5$ hour video of the vehicle detector ran on our test set. This may be found at youtu.be/GJ0cZBkHoHc. A short video of our lane detector may also be found online at youtu.be/\texttt{\_\_}f5pqqp6aM. In these videos, we evaluate the detector on every frame independently and display the raw detections, without the use of any Kalman filters or road models. The red locations in the video correspond to the mask detectors that are activated. This network was only trained on the rear view of cars traveling in the same direction, which is why cars across the highway barrier are commonly missed.

We have open sourced the code for the vehicle and lane detector online at github.com/brodyh/caffe. Our repository was forked from the original Caffe code base from the BVLC group~\cite{jia-2014}.

\section{Conclusion}
By using Camera, Lidar, Radar, and GPS we built a highway data set consisting of $17$ thousand image frames with vehicle bounding boxes and over $616$ thousand image frames with lane annotations. We then trained on this data using a CNN architecture capable of detecting all lanes and cars in a single forward pass. Using a single GTX 780 Ti our system runs at $44$Hz, which is more than adequate for real-time use. Our results show existing CNN algorithms are capable of good performance in highway lane and vehicle detection. Future work will focus on acquiring frame level annotations that will allow us to develop new neural networks capable of using temporal information across frames.

\section*{Acknowledgment}

This research was funded in part by Nissan who generously donated the car used for data collection. We thank our colleagues Yuta Yoshihata from Nissan who provided technical support and expertise on vehicles that assisted the research. In addition, the authors would like to thank the author of Overfeat, Pierre Sermanet, for their helpful suggestions on image detection.


\begin{thebibliography}{1}

\bibitem{sermanet-2013}  
Sermanet, Pierre, et al. "Overfeat: Integrated recognition, localization and detection using convolutional networks." arXiv preprint arXiv:1312.6229 (2013).

\bibitem{rothengatter-1997}
Rothengatter, Talib Ed, and Enrique Carbonell Ed Vaya. "Traffic and transport psychology: Theory and application." International Conference of Traffic and Transport Psychology, May, 1996, Valencia, Spain. Pergamon/Elsevier Science Inc, 1997.

\bibitem{cho-2014}
Cho, Hyunggi, et al. "A multi-sensor fusion system for moving object detection and tracking in urban driving environments." Robotics and Automation (ICRA), 2014 IEEE International Conference on. IEEE, 2014.

\bibitem{held-2012}
Held, David, Jesse Levinson, and Sebastian Thrun. "A probabilistic framework for car detection in images using context and scale." Robotics and Automation (ICRA), 2012 IEEE International Conference on. IEEE, 2012.

\bibitem{levinson-2011}
Levinson, Jesse, et al. "Towards fully autonomous driving: systems and algorithms." Intelligent Vehicles Symposium, 2011. 


\bibitem{carafii-2012}
Caraffi, Claudio, et al. "A system for real-time detection and tracking of vehicles from a single car-mounted camera." Intelligent Transportation Systems (ITSC), 2012 15th International IEEE Conference on. IEEE, 2012.

\bibitem{jazayeri-2011}
Jazayeri, Amirali, et al. "Vehicle detection and tracking in car video based on motion model." Intelligent Transportation Systems, IEEE Transactions on 12.2 (2011): 583-595.

\bibitem{opencv}
Bradski, Gary. "The opencv library." Doctor Dobbs Journal 25.11 (2000): 120-126.

\bibitem{krizhevsky-2012}
Krizhevsky, Alex, Ilya Sutskever, and Geoffrey E. Hinton. "Imagenet classification with deep convolutional neural networks." Advances in neural information processing systems. 2012.

\bibitem{szegedy-2013}
Szegedy, Christian, Alexander Toshev, and Dumitru Erhan. "Deep neural networks for object detection." Advances in Neural Information Processing Systems. 2013.

\bibitem{sutskever-2013}
Sutskever, Ilya, et al. "On the importance of initialization and momentum in deep learning." Proceedings of the 30th International Conference on Machine Learning (ICML-13). 2013.

\bibitem{eigen-2014}
Eigen, David, Christian Puhrsch, and Rob Fergus. "Depth map prediction from a single image using a multi-scale deep network." Advances in Neural Information Processing Systems. 2014.

\bibitem{felzenszwalb-2010}
Felzenszwalb, Pedro F., et al. "Object detection with discriminatively trained part-based models." Pattern Analysis and Machine Intelligence, IEEE Transactions on 32.9 (2010): 1627-1645.

\bibitem{szegedy-2014}
Szegedy, Christian, et al. "Scalable, High-Quality Object Detection." arXiv preprint arXiv:1412.1441 (2014).

\bibitem{girshick-2014}
Girshick, Ross, et al. "Rich feature hierarchies for accurate object detection and semantic segmentation." Computer Vision and Pattern Recognition (CVPR), 2014 IEEE Conference on. IEEE, 2014.

\bibitem{uijlings-2013}
Uijlings, Jasper RR, et al. "Selective search for object recognition." International journal of computer vision 104.2 (2013): 154-171.

\bibitem{szegedy-2014-inception}
Szegedy, Christian, et al. "Going deeper with convolutions." arXiv preprint arXiv:1409.4842 (2014).

\bibitem{he-2015}
He, Kaiming, et al. "Delving Deep into Rectifiers: Surpassing Human-Level Performance on ImageNet Classification." arXiv preprint arXiv:1502.01852 (2015).

\bibitem{simonyan-2014}
Simonyan, Karen, and Andrew Zisserman. "Very deep convolutional networks for large-scale image recognition." arXiv preprint arXiv:1409.1556 (2014).

\bibitem{jia-2014}
Jia, Yangqing, et al. "Caffe: Convolutional architecture for fast feature embedding." Proceedings of the ACM International Conference on Multimedia. ACM, 2014.

\end{thebibliography}
\end{document}